\documentclass[runningheads]{llncs}
\usepackage{graphicx}
\usepackage{soul, color}
\usepackage{subcaption}
\usepackage[perpage]{footmisc}

\begin{document}
\title{AutoFB: Automating Fetal Biometry Estimation from Standard Ultrasound Planes}
\titlerunning{AutoFB: Automating Fetal Biometry Estimation}

\author{Sophia Bano \inst{1,2} \and
Brian Dromey \inst{1,3} \and 
Francisco Vasconcelos \inst{1,2} \and
Raffaele Napolitano \inst{3} \and 
Anna L. David \inst{3,4} \and 
Donald M. Peebles \inst{3,4} \and
Danail Stoyanov \inst{1,2}}

%

\institute{Wellcome/EPSRC Centre for Interventional and Surgical Sciences(WEISS), University College London, London, UK  \\
\email{sophia.bano@ucl.ac.uk} \and
Department of Computer Science, University College London, London, UK \and
Elizabeth Garrett Anderson Institute for Women's Health, University College London, London, UK \and
NIHR University College London Hospitals Biomedical Research Centre, London, UK}

\authorrunning{S. Bano et al.}

\maketitle

\begin{abstract}
During pregnancy, ultrasound examination in the second trimester can assess fetal size according to standardized charts. To achieve a reproducible and accurate measurement, a sonographer needs to identify three standard 2D planes of the fetal anatomy (head, abdomen, femur) and manually mark the key anatomical landmarks on the image for accurate biometry and fetal weight estimation. This can be a time-consuming operator-dependent task, especially for a trainee sonographer. Computer-assisted techniques can help in automating the fetal biometry computation process. In this paper, we present a unified automated framework for estimating all measurements needed for the fetal weight assessment. The proposed framework semantically segments the key fetal anatomies using state-of-the-art segmentation models, followed by region fitting and scale recovery for the biometry estimation. We present an ablation study of segmentation algorithms to show their robustness through 4-fold cross-validation on a dataset of 349 ultrasound standard plane images from 42 pregnancies. Moreover, we show that the network with the best segmentation performance tends to be more accurate for biometry estimation. Furthermore, we demonstrate that the error between clinically measured and predicted fetal biometry is lower than the permissible error during routine clinical measurements. 

\end{abstract}
\section{Introduction}
\begin{figure}[t!]
\centering
	\begin{subfigure}[b]{0.33\textwidth}
		\centering
		\includegraphics[width=1.0\textwidth]{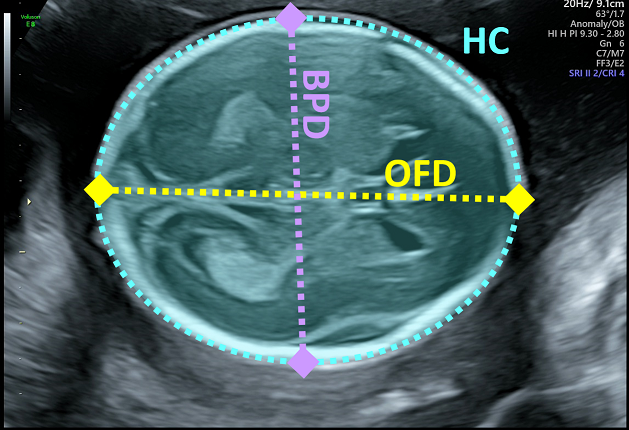}
	\end{subfigure}	
	\begin{subfigure}[b]{0.30\textwidth}
	\centering
	\includegraphics[width=1.0\textwidth]{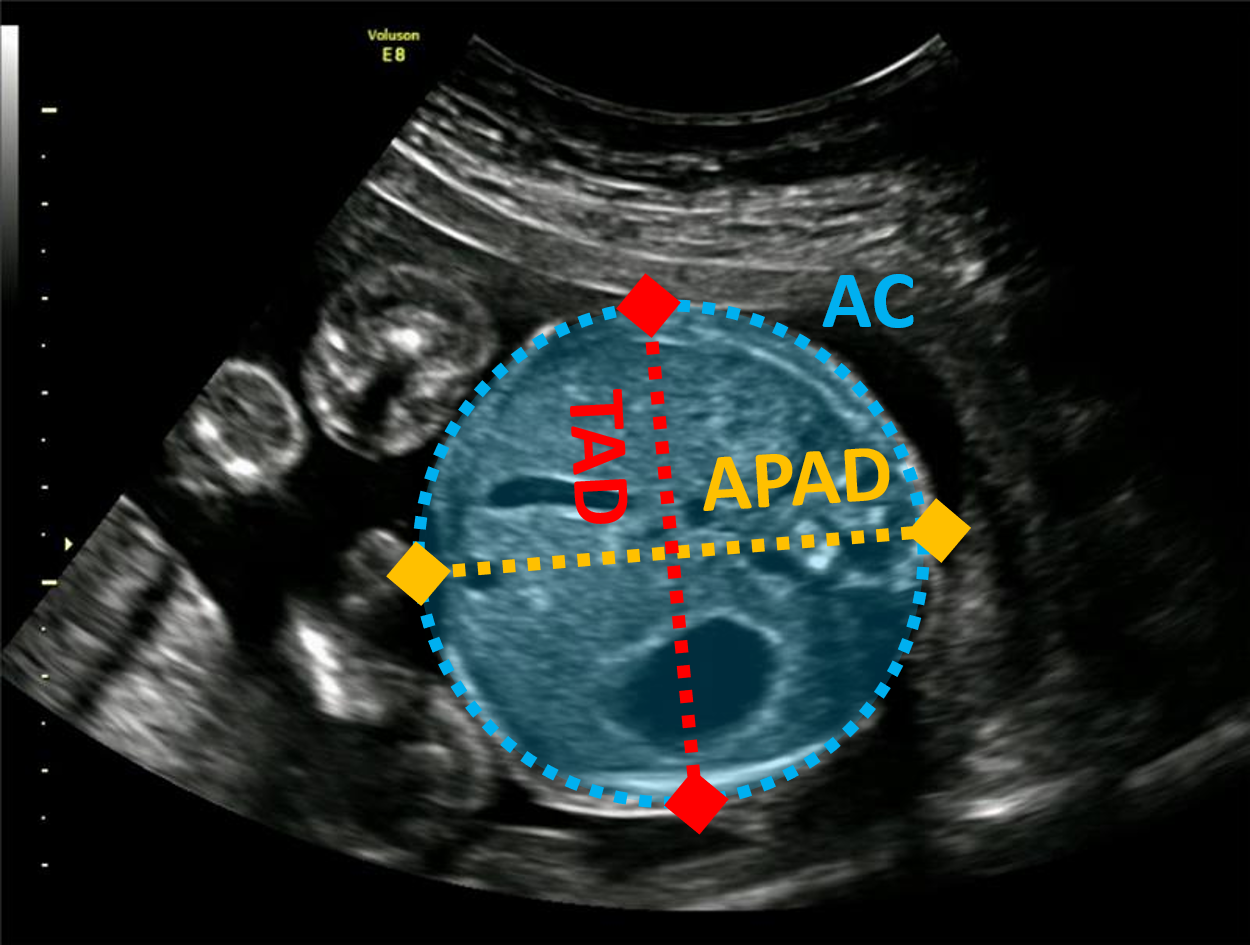}
    \end{subfigure}
	\begin{subfigure}[b]{0.26\textwidth}
		\centering
		\includegraphics[width=1.0\textwidth]{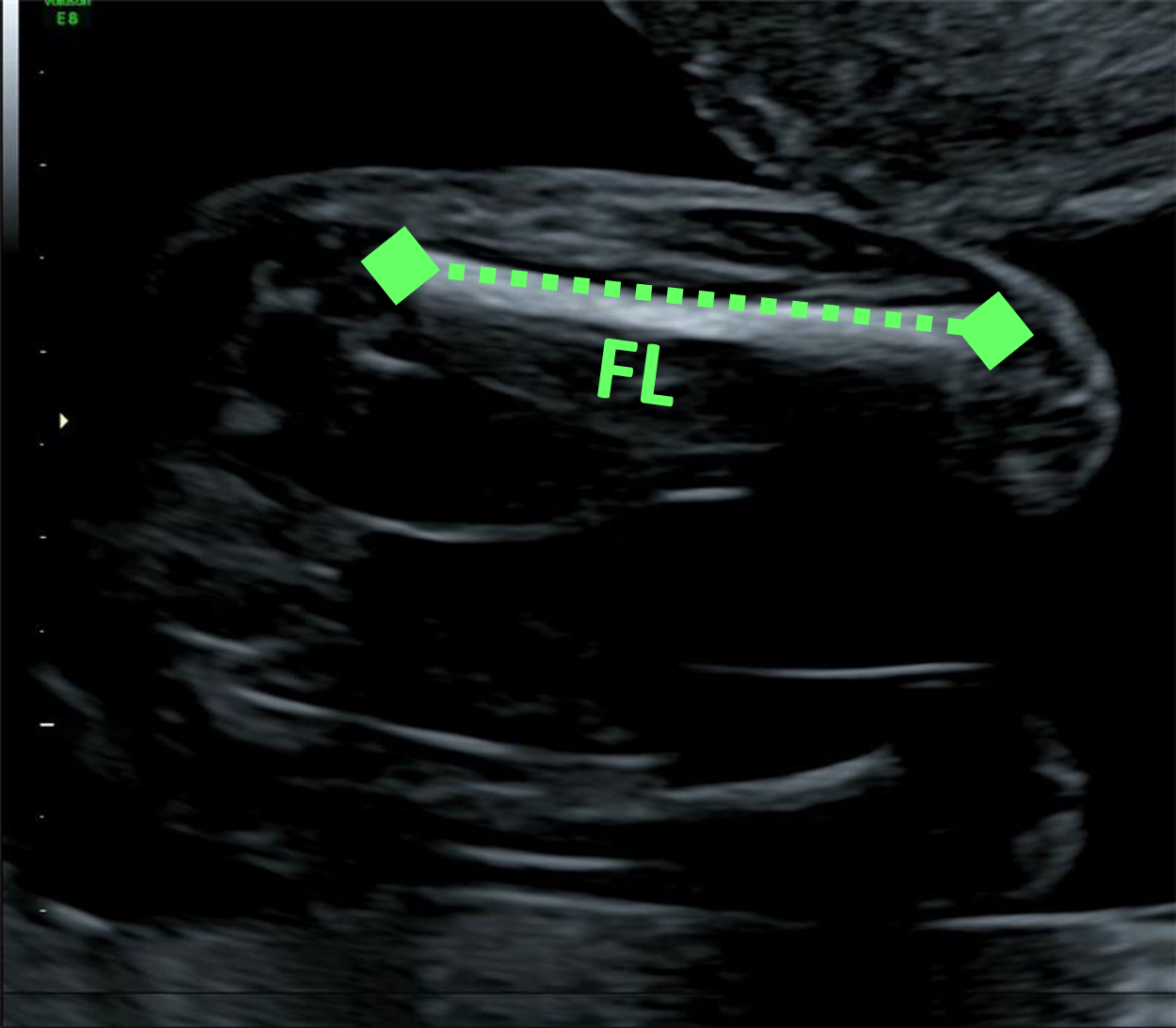}
	\end{subfigure}	
	\caption{Fetal biometry from transventricular plane in the head (left), transabdominal plane in the abdomen (middle) and femur plane (right).}
    \label{fig:fb_illustration}
\end{figure}

There is little global consensus on how to train, assess and evaluate skills in prenatal second trimester ultrasound (US) screening. Recommended assessment and quality control metrics vary across countries and institutions \cite{dromey2020dimensionless}. Despite this, standardized US planes and metrics to assess fetal size are well established \cite{salomon2019isuog}. In particular, fetal weight estimation is routinely used to assess fetal well-being, both in terms of its absolute value and its growth trajectory during pregnancy. Fetal wellbeing is considered by obstetricians for scheduling birth and by neonatologists when counselling parents on likely outcomes for their baby. There are three key structures and corresponding anatomical planes which are used for the estimation of fetal weight (Fig.~\ref{fig:fb_illustration}). These are the transventricular plane to measure the head, the transabdominal plane to measure the abdomen and the femur length plane to measure the leg skeletal size. The acquisition of these standard planes is subject to intraoperator and interoperator variabilities~\cite{sarris2012intra} which introduces some degree of uncertainty in the clinically obtained weight measurements and consequently requires a degree of caution when clinicians are interpreting fetal growth reports. Sonography expertise has a significant impact on minimizing variability of image quality and fetal biometry \cite{dromey2020dimensionless}. Consequently, training and competence assessment are of great importance to ensure effective, reproducible and safe clinical practice. Automating fetal biometry on the standardized planes can help in minimizing the variability, specially in the case of less experienced sonographers and may also serve as expert for trainees. 

There is extensive work on segmentation of anatomical structures in 
standard US planes, specifically those concerning second and third trimester screening \cite{rueda2013evaluation}. These techniques can support automated fetal biometry, including measurements on the head \cite{zhang2017automatic,van2018automated,li2017automatic,sobhaninia2019fetal,khan2017automatic,chen2020automatic}, femur \cite{khan2017automatic,hermawati2019automatic}, and abdominal section \cite{khan2016automatic}. These methods, however, rely on prior knowledge of which measurement to perform on a given image. A fully automated biometry system should both identify which standard plane is being imaged and whether it is of sufficient quality to perform the relevant measurements. Automatic image quality assessment has been investigated, including adequate magnification, symmetry and the visibility of relevant anatomical structures within the image \cite{lin2019multi,khan2017automatic}. Such methods together with classification of standard planes \cite{baumgartner2017sononet} can be used to extract appropriate planes for fetal biometry from US video or image collections \cite{gao2020label}. Alternative approaches involve obtaining standard planes from 3D US volumes \cite{grandjean2018artificial}, in which the extracted planes approach those of an experienced sonographer but results are so far limited to the fetal head measurements. Standard plane classification has also been further developed to provide active guidance during freehand operation \cite{droste2020automatic}.

In this paper, we propose performing all the relevant measurements for fetal weight estimation within a unified automated system, which is our main contribution. The proposed AutoFB framework involves classifying the three standard planes and segmenting the head, abdomen and femur. This is followed by the extraction of the following measurements: biparietal diameter (BPD), occipito‐frontal diameter (OFD), head circumference (HC), transverse abdominal diameter (TAD), anterior‐posterior abdominal diameter (APAD), abdominal circumference (AC), and femur length (FL). We achieve this by training a multi-class segmentation neural network that automatically identifies and segments the relevant anatomy structures within any of the three standard planes. The corresponding biometry is then extracted by applying scale recovery and using ellipse fitting (head or abdomen) and bounding box fitting (femur). To the best of our knowledge, AutoFB is the first framework to automate fetal biometry estimation from all three standard planes. We demonstrate the robustness of AutoFB by experimenting using real clinical US data and validate both inferred segmentation and estimated biometry. The clinical data used for the validation contains 346 2D US planes from 42 pregnancies. AutoFB is of high clinical relevance as it will enable automating biometry, a task currently affected by high inter-operator variability~\cite{dromey2020dimensionless} due to manual selection and measurement of the relevant US plane.  

\section{Fetal Biometry}
To clinically measure fetal size and weight during a fetal US, the sonographer navigates the US probe to localize a view of each of the three standard planes. While this task is subject to operator variability, there are established guidelines on which features should be visible within each standard plane \cite{cavallaro2018quality}. They must then lock the display and manually place calipers on key landmarks from which biometric measurements are extracted. 
The BPD and  OFD measurements are required for the HC measurement on the transventricular plane (Fig.~\ref{fig:fb_illustration}(left)). The TAD and APAD are required for the AC measurement on the transabdominal plane (Fig.~\ref{fig:fb_illustration}(middle)). HC and AC are then computed using, $\pi(d_1+d_2)/2$, where $d_1$ and $d_2$ are the BPD and OFD in the case of head and TAD and APAD in the case of abdomen measurements. Alternatively, an ellipse fitting function is available in some US machines and can be used for head and abdominal measurements, however, its usage largely depends on operator choice or established practice within a specific clinical site. This feature is not routinely used in the context of data acquired and presented within this work.
To measure the femur length (FL), the extreme lateral edges including both of the epiphyses must be visualized and measured along the long axis of the femur (Fig.~\ref{fig:fb_illustration}(right)). \looseness=-1

\section{Methodology}
\begin{figure}[t!]
\centering
  \includegraphics[width=0.9\textwidth]{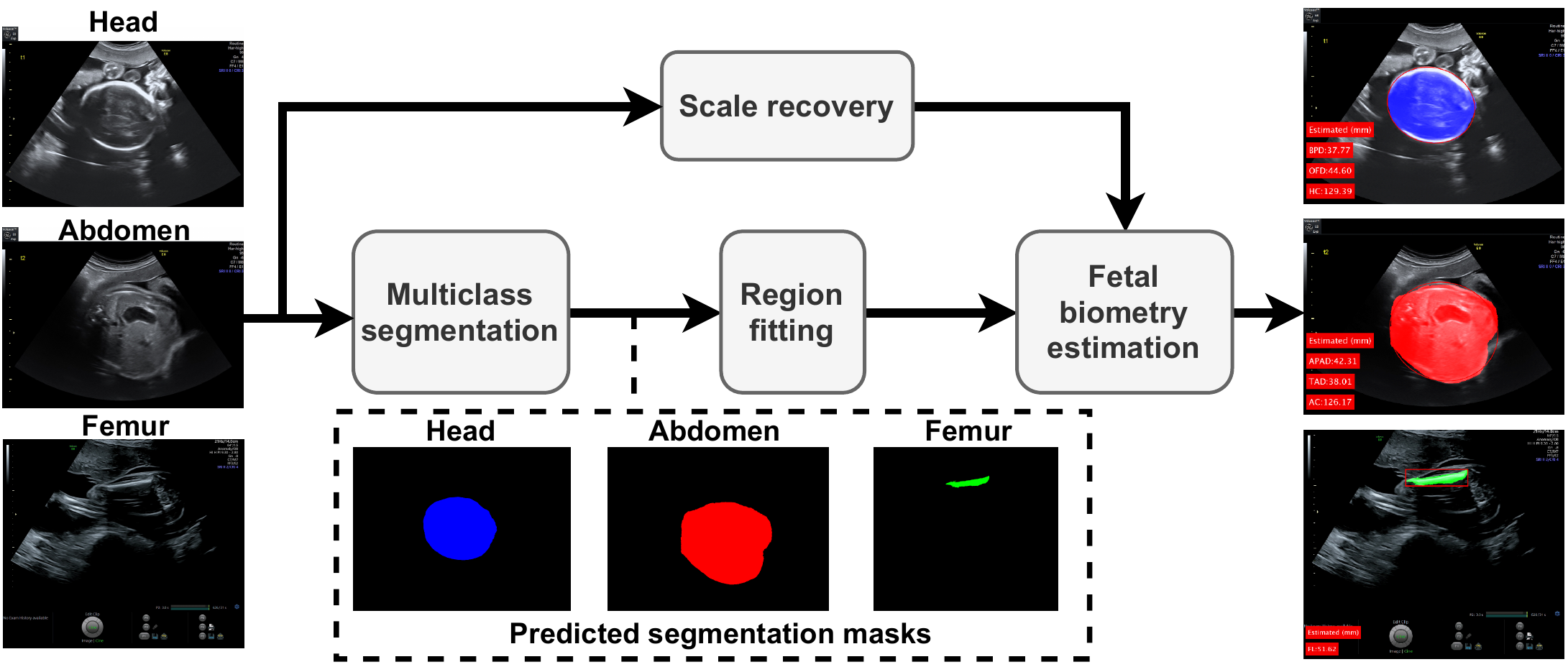}
  \caption{Overview of the AutoFB framework. Given a US standard plane, AutoFB performs multi-class segmentation for plane detection and anatomy segmentation, followed by shape fitting and scale recovery for biometry estimation.}
  \label{fig:block_diagram}
\end{figure}

An overview of the proposed framework is presented in Fig.~\ref{fig:block_diagram}. The framework jointly performs the 3-plane detection and anatomy segmentation by training state-of-the-art segmentation models for identifying the head, abdomen and femur anatomies and selecting the best performing architecture (Sec.~\ref{sec:img_seg}). This is followed by shape fitting on the segmented regions, automated image scale retrieval and biometry estimation in millimetres units (Sec.~\ref{sec:FBE}). 

\subsection{Multi-class Image Segmentation}
\label{sec:img_seg}
In order to build a unified system, we define our problem as semantic segmentation between 4 specific classes: head, abdomen, femur, and background. With groundtruth data, each standard plane will only contain background and one of the other 3 classes. We experimented with two state-of-the-art image segmentation models, namely, U-Net~\cite{ronneberger2015u} and Deeplabv3+~\cite{chen2018encoder}. U-Net can be regarded as the most commonly used architecture for biomedical image segmentation and is recommended when the training data is limited. Deeplabv3+ has achieved state-of-the art performance on large-scale semantic segmentation datasets (PASCAL VOC 2012). Both U-Net and Deeplabv3+ are encoder-decoder networks, where U-Net is a special case in which the decoder component is connected with the encoder through skip connections and is not decoupled from the encoder. We briefly introduce these architectures and refer the reader to \cite{ronneberger2015u,chen2018encoder} for specific details. \looseness = -1

U-Net is a type of fully convolutional network which consists of a contraction path and an expansion path. The contraction path can be a pretrained encoder which captures the context while limiting the feature map size. The expansion path is a symmetric decoder network which also performs up-sampling to recover the segmentation map size. The encoder and decoder paths are connected through skip connections for sharing localization information. We used the ResNet50~\cite{he2016deep} as the encoder architecture for U-Net. We also experimented with Mobilenetv2~\cite{sandler2018mobilenetv2} to have a fair comparison of the two segmentation architectures under analysis.  
Deeplabv3+~\cite{chen2018encoder} uses several parallel atrous convolutions (also known as dilated convolutions) with different rates to capture the contextual information at multiple scales without losing image resolution. This approach is referred to as Atrous Spatial Pyramid Pooling. Moreover, Deeplabv3+ recovers the detailed object boundaries through a simple yet effective decoder module~\cite{chen2018encoder}. 
We used MobileNetv2~\cite{sandler2018mobilenetv2} instead of Xception model (that was used in~\cite{chen2018encoder}) as the backbone for DeeplabV3+ as MobileNetv2 backbone is both light-weight and effective. 

We use cross entropy (CE) as loss function. From Table~\ref{tab:dataset}, we can observe that the data is highly imbalanced, with the femur class having much fewer samples compared to head, abdomen and background classes due to its comparatively small segmentation area. To handle this issue, we also use weighted CE (wCE) where given the total number of pixels per class, $[c_i]_{i}^{4}$, weight $w_i$ for the $i^{th}$ class is given by, $w_i = \frac{max([c_i]_{i}^{4})}{c_i}$. The obtained results are discussed in Sec.~\ref{sec:results}. 

\subsection{Fetal Biometry Estimation}
\label{sec:FBE}

Different standard planes require different biometry measurements, and therefore the first step is to detect and localize the segmented region. This is defined as the largest segmented area predicted by the networks described in the previous section. We later show experimentally that this strategy correctly identifies all planes in our test data. It is known a priori that the head and abdomen are elliptical while the femur is oblong (Fig.~\ref{fig:fb_illustration}). Thus, ellipse fitting is performed on the segmented head and abdomen masks through shape contour extraction and applying direct least square to fit an ellipse~\cite{fitzgibbon1999direct}, where the major and minor axes of the fitted ellipse represent BPD and OFD for the head and TAD and APAD for the abdomen, respectively. These are in turn used to calculate the circumference of the fitted ellipses, providing HC and AC measurements. On the femur plane, a horizontal bounding box with zero orientation is fitted on the segmented mask, where the length of its diagonal gives the FL estimate. A femur is not necessarily aligned to the horizontal/vertical axis, hence the use of bounding box diagonal as FL always holds. Finally, lengths in pixels are scaled to millimetres to obtain results that are directly comparable to clinically measured biometry. 

While the metric scale of the US images (in px/mm) is usually trivial to obtain during operation, the automatic extraction of this parameter from retrospectively acquired data proved useful to fully automate the hundreds of measurements obtained in this work. Obtaining US scale is always system-dependent because it must be extracted either from the visual interface of the US machine or from the raw data, which requires access to a proprietary API. We use visual interface for scale recovery since we did not have access to the raw data. To obtain the scale, we exploit the consistent interface of the US machine used to acquire our dataset (GE Voluson), namely the caliper visible on the left-hand side of the US images. The ruler markers are detected with simple template matching and their smallest interval (can be either 5mm or 10mm) is determined from the relative size of the markers. The same template matching approach is easy to deploy on systems other than GE Voluson since all medical grade US machines have a similar ruler available. 

\section{Dataset and Experimental Setup}
The data collection process has been reviewed and approved by the local research ethics committee under the title; "Fetal US and fetal monitoring technologies to improve prenatal diagnosis and therapy for fetal abnormality and maternal and fetal perinatal outcome", IRAS ID 230125. Patients attending University College London Hospital for US examination were enrolled and pseudo-anonomyzed by the clinical research staff. Each patient gave written consent. For the purpose of anonymizing, transferring and storing data a customized version of XNAT 1.6 was used. The complete image library from each US was transferred to the research database. The hospital protocol undertakes US screening in accordance with a National Fetal Anomaly Screening Program~\footnote{\url{ https://tinyurl.com/NHSFetalAnomalyScreeningHB}}. Each saved image represents an image of diagnostic quality. The US images saved by the operator were considered to be the optimal image for that scan given the limitations of fetal lie and stage of gestation.  The measurement calipers were applied by the US operator and in most cases,  the image with and without the measurement calipers were saved. A subset of images relevant to fetal biometry were extracted from the database by a clinical research fellow. A total of 346 images were included from 42 pregnancies. Each image in the set of data was classified as AC, HC or FL. The VIA annotation tool~\cite{dutta2019vgg} was used to manually annotate the head, abdomen or femur within each image for the segmentation task. 
The obtained fully anonymized standard US plane images have large intra-class variability. For example, in some cases the femur is well aligned to the horizontal plane while in other cases the angle of sonnation is wider and the level of magnification is less. Although operators followed a standard protocol to capture a good quality image including all necessary anatomical details, some images have relatively poor  contrast and dark patches. These are secondary to the technical limitations of US and maternal body habitus. Often unavoidable, the heterogeneity of the data set introduces  challenges for the segmentation task. 
\begin{table}[t!]
	\centering
	\caption{Total number of sample in each segmentation class and in each cross-validation fold and average pixels per class per frames.}
	\label{tab:dataset}
	\resizebox{1.0\textwidth}{!}{
	\footnotesize
	\begin{tabular}{|l|c|c|c|c|c|l|l|r|}
		\hline 
	 &\textbf{All images} &\textbf{Fold 1} &\textbf{Fold 2} &\textbf{Fold 3} &\textbf{Fold 4} & &\multicolumn{2}{c|}{\textbf{Avg. pixels per class }}\\ \cline{1-6} 
		\textbf{Total subjects} &42 &10 &9  &12 &11 & & \multicolumn{2}{c|}{\textbf{per frame}} \\ \hline
		\textbf{Total Images} &346 &87 &86  &89 &84 & &\textbf{Background} &816239\\ \hline
		\textbf{Head} &135 &26 &44  &29 &36 & &\textbf{Head} &74127\\ \hline
		\textbf{Abdomen} &103  &32 &22 &26 &23 & &\textbf{Abdomen} &44691\\ \hline
		\textbf{Femur}  &108 &29  &20 &34 &25 & &\textbf{Femur} &3833\\ \hline
    \end{tabular}
   }
\end{table}

The acquired data from 42 fetuses (346 US images) is divided into 4 folds, used for testing the robustness of the segmentation networks, such that each fold contains at least 80 images and all US images originating from a single fetus are only included in a particular fold. Hence, the data in a fold is unseen for all other folds (as mentioned in Table~\ref{tab:dataset}). Mean Intersection over Union (mIoU) is used for evaluating the segmentation models, and absolute error between the clinically measured and automatically predicted fetal biometry is used for evaluating the proposed AutoFB. All images are of varying sizes (resolution) as they were cropped to remove any identifiable information. Therefore, we resized all images to $1024\times1024$ pixel resolution before model training. Data augmentation is applied by introducing random scale, rotation, shift, flipping, brightness and contrast changes before obtaining an image crop of size $512\times512$ pixel at a random location which is used as the input for training the segmentation network. Data augmentation helped in avoiding model over-fitting. An initial learning rate $10e^{-3}$ with a step decay by a factor of $1/10$ at $75^{th}$ and $150^{th}$ is used with the ADAM optimizer. The model is trained for 600 epochs with early stopping based on the criteria of no improvement of the training set with patience of 50 epoch is used. The weights that captured the best performance on the training data are used to evaluate the segmentation model on the holdout fold. The segmentation networks are implemented in PyTorch and trained using a single Tesla V100-DGXS-32GB GPU of an NVIDIA DGX-station. 

\begin{table}[t!]
	\centering
	\caption{Four-fold cross-validation results showing comparison of Deeplabv3+ and UNet having different configurations. Mean and standard deviation of mIoU across all folds is reported. Key: BG- background; H - head; A - abdomen; F - femur; CE - cross entropy; wCE - weighted cross entropy; MNv2 - Mobilenetv2.}
	\label{tab:quan_eval}
	\resizebox{1.0\textwidth}{!}{
	\footnotesize
	\begin{tabular}{|l|c|c|c|c|c|}
		\hline 
	 \textbf{Method} &\textbf{mIoU} &\textbf{mIoU-BG} &\textbf{mIoU-H} &\textbf{mIoU-A} &\textbf{mIoU-F}\\ \hline \hline
	 \textbf{Deeplabv3+ (MNv2-CE)} &0.87$\pm$0.02	&0.95$\pm$0.02	&0.93$\pm$0.02	&0.89$\pm$0.03	&0.61$\pm$0.03 \\ \hline
	 \textbf{Deeplabv3+ (MNv2-wCE)} &\textbf{0.88}$\pm$\textbf{0.01} &\textbf{0.95}$\pm$\textbf{0.01}	&\textbf{0.93}$\pm$\textbf{0.02}	&\textbf{0.89}$\pm$\textbf{0.02}	&\textbf{0.61}$\pm$\textbf{0.02} \\ \hline \hline
	 \textbf{UNet (MNv2-CE)} &$0.82\pm0.05$	&$0.93\pm0.03$	&$0.89\pm0.05$	&$0.85\pm0.05$	&$0.56\pm0.03$ \\ \hline
	 \textbf{UNet (MNv2-wCE)} &$0.86\pm0.01$	&$0.94\pm0.01$	&$0.91\pm0.02$	&$0.86\pm0.02$	&$0.58\pm0.01$ \\ \hline \hline
	 \textbf{UNet (Resnet-CE)}  &$0.75\pm0.06$	&$0.88\pm0.05$	&$0.84\pm0.07$	&$0.77\pm0.05$	&$0.53\pm0.03$ \\ \hline
	 \textbf{UNet (Resnet-wCE)} &$0.78\pm0.04$	&$0.87\pm0.03$	&$0.83\pm0.04$	&$0.75\pm0.06$	&$0.53\pm0.02$ \\ \hline
    \end{tabular}
   }
\end{table}

\section{Results and Discussion}
\label{sec:results}
We perform comparison of the Deeplabv3+ and U-Net having two commonly used backbones and used both CE and wCE losses (refer to Sec.~\ref{sec:img_seg}). The quantitative comparison using 4-fold cross-validation is presented in Table~\ref{tab:quan_eval}. 
Both configurations of Deeplabv3+ are comparable (overall mIoU = 0.88) though the standard deviation is lower when wCE is used. Deeplabv3+ also outperformed the UNet configurations. The effect of introducing wCE loss for handling class imbalance problem is more evident from the different UNet configurations. Mobilenetv2 backbone, which has significantly less number of network parameters (3.5M), showed superior performance than the Resnet50 (26M parameters) backbone. Selecting an efficient and robust backbone architecture is essential and can significantly improve the overall segmentation network performance. From Table~\ref{tab:quan_eval}, we can observe that mIoU-F is particularly low compared to the mIoU-BG, mIoU-H and mIoU-A. This is because (1) the number of per-pixel samples in the femur class are very small (Table.~\ref{tab:dataset}); (2) a small error in predicted segmentation vs the ground-truth results in a significantly low IoU value when the object size is small; (3) of large intraclass variability.

\begin{figure}[t!]
	\begin{subfigure}[b]{0.40\textwidth}
		\centering
        \includegraphics[width=1\textwidth]{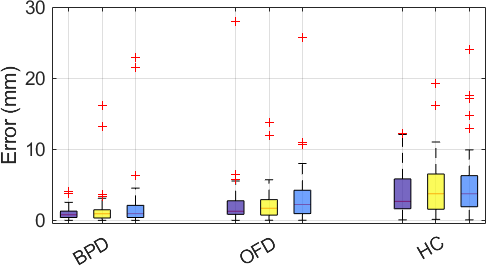}	\caption{Head}	
	\end{subfigure}	
	\hfill
	\begin{subfigure}[b]{0.40\textwidth}
		\centering
		\includegraphics[width=1\textwidth]{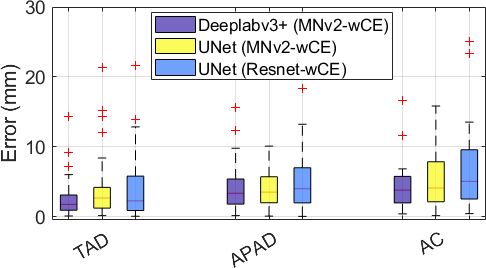}
		\caption{Abdomen}	
	\end{subfigure}	
	\hfill
	\begin{subfigure}[b]{0.135\textwidth}
		\centering
        \includegraphics[width=1\textwidth]{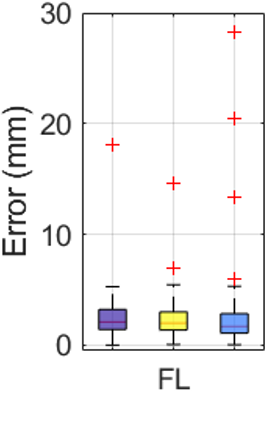}
		\caption{Femur}	
	\end{subfigure}	
\caption{Boxplots showing the comparison between the best performing models and the absolute error between the clinically measured and predicted fetal biometry.}
\label{fig:Err_plots}
\end{figure}

\begin{figure*}[t!]
	\centering
	\footnotesize
	\resizebox{1\columnwidth}{!}{
	\begin{tabular}{c c c c c c}
	&\multicolumn{1}{c}{Input image} & \multicolumn{1}{c}{Ground-truth} &\multicolumn{1}{c}{DLv3+ (MNv2-wCE)} &\multicolumn{1}{c}{UNet (MNv2-wCE)} &\multicolumn{1}{c}{UNet (Resnet-wCE)}  \\ 
	\rotatebox[origin=c]{90}{HC} &\parbox[c]{2.9cm}{\centering \includegraphics[width=2.9cm]{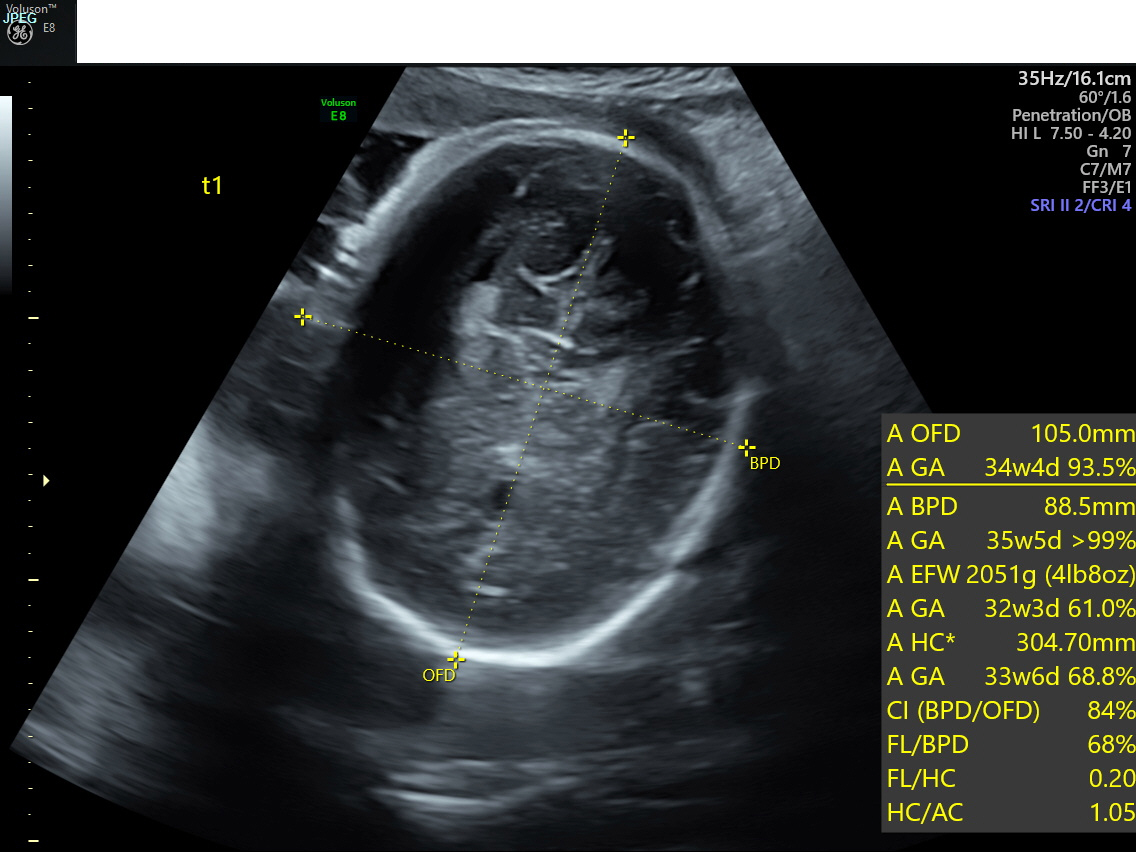}} 
	    &\parbox[c]{2.9cm}{\centering	\includegraphics[width=2.9cm]{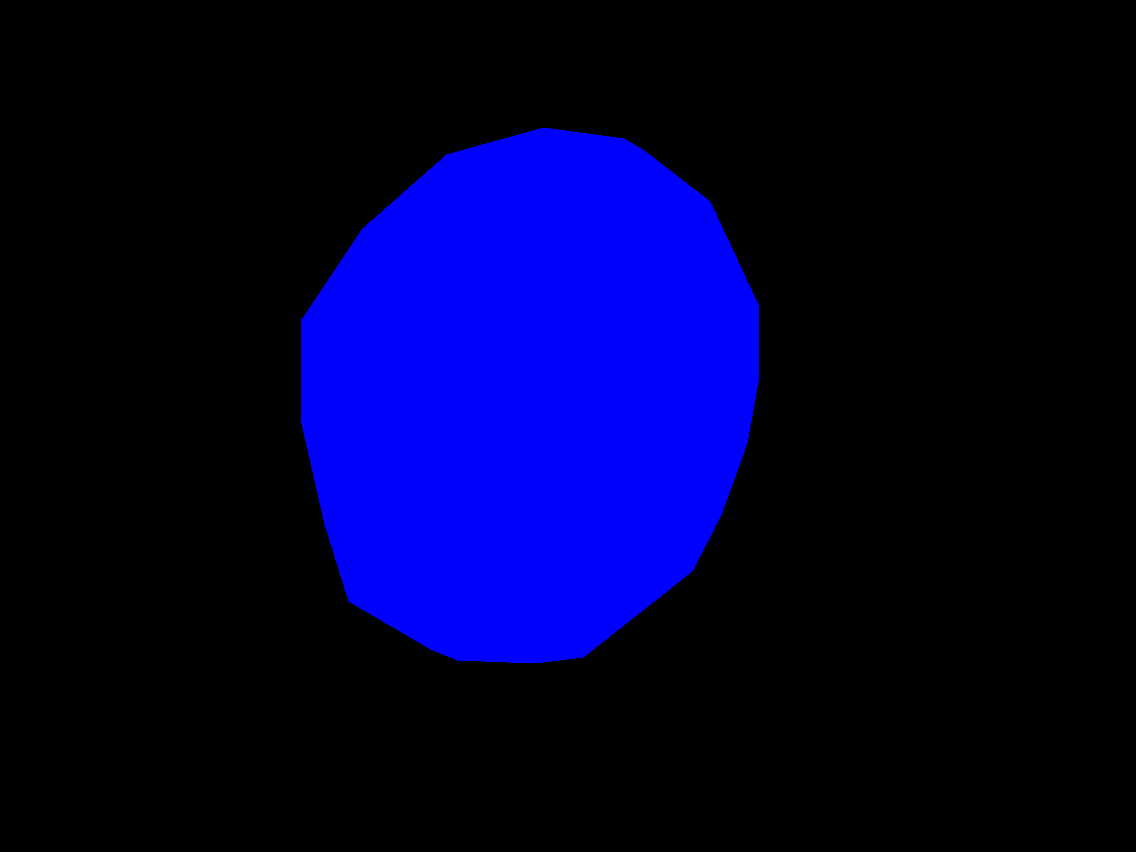}} 
	    &\parbox[c]{2.9cm}{\centering \includegraphics[width=2.9cm]{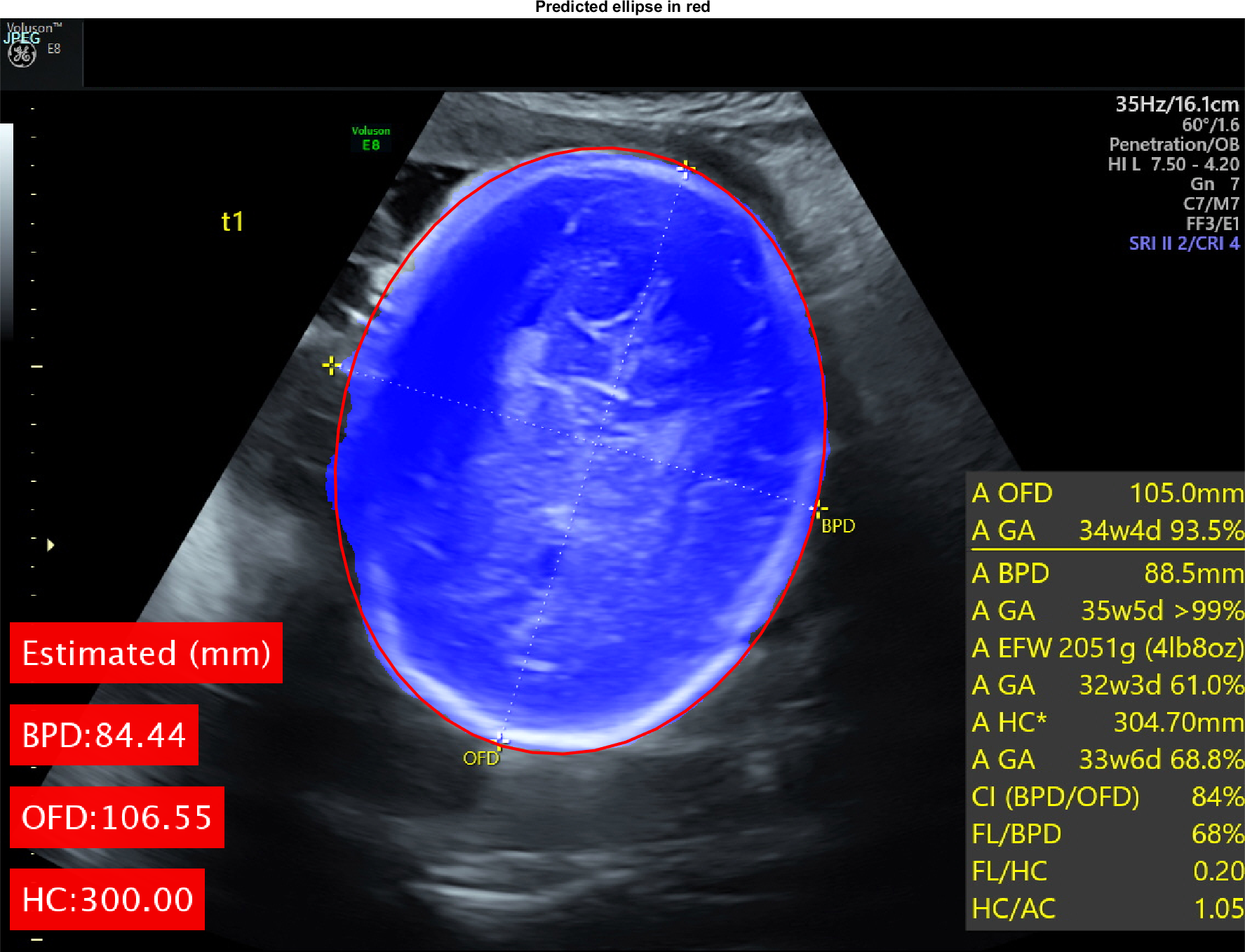}} 
	    &\parbox[c]{2.9cm}{\centering \includegraphics[width=2.9cm]{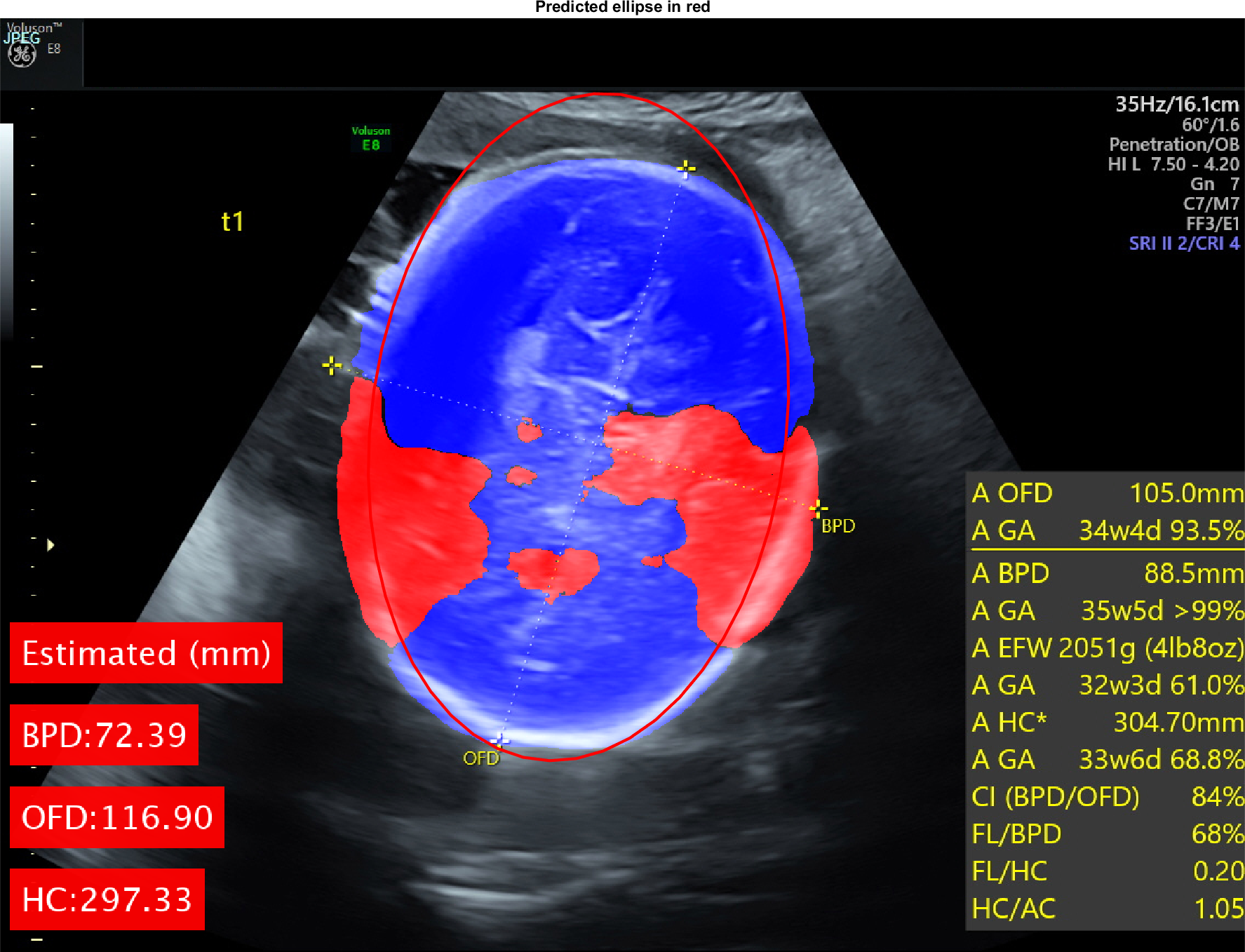}}
	    &\parbox[c]{2.9cm}{\centering \includegraphics[width=2.9cm]{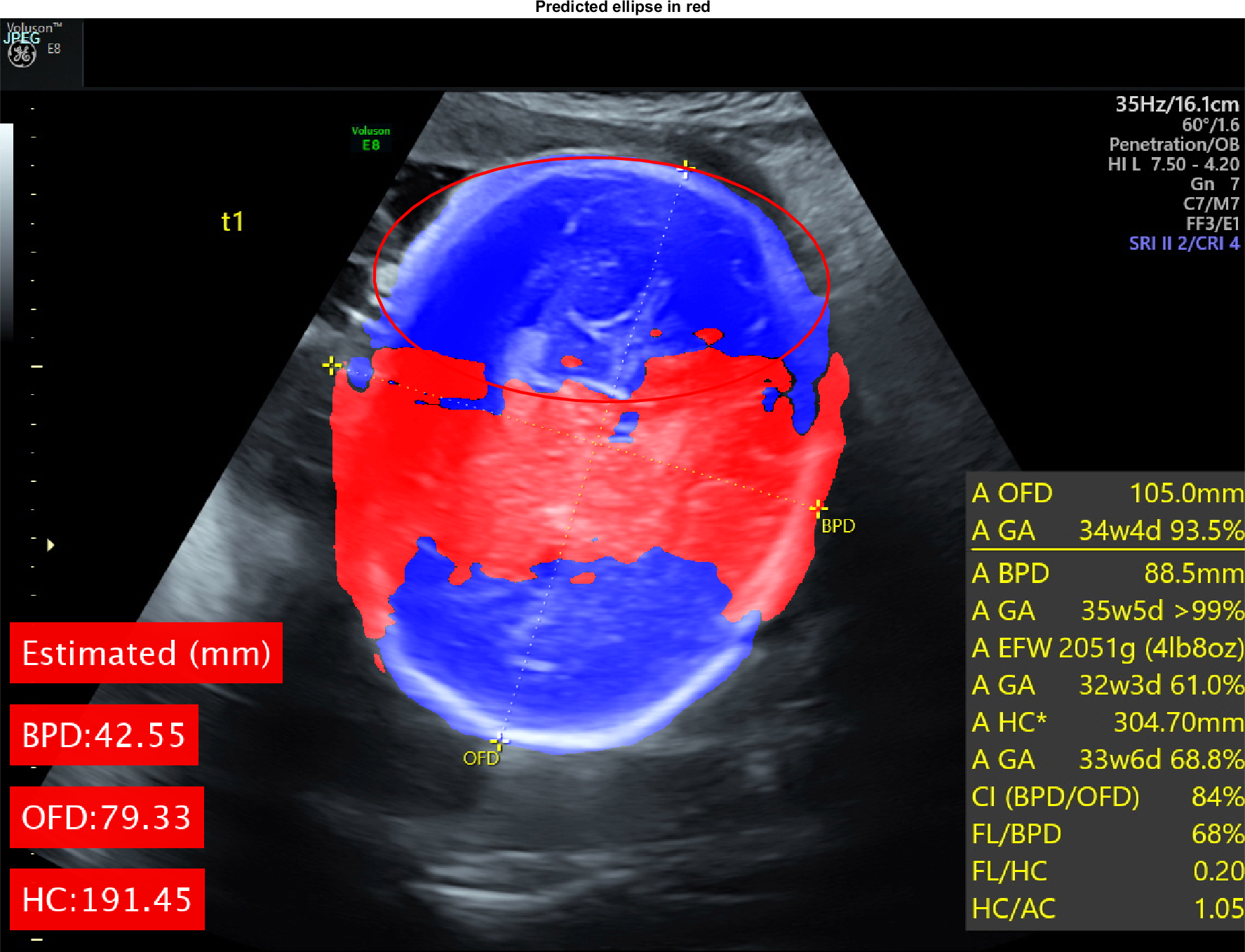}} 
	    \\
	 \rotatebox[origin=c]{90}{AC} &\parbox[c]{2.9cm}{\centering \includegraphics[width=2.9cm]{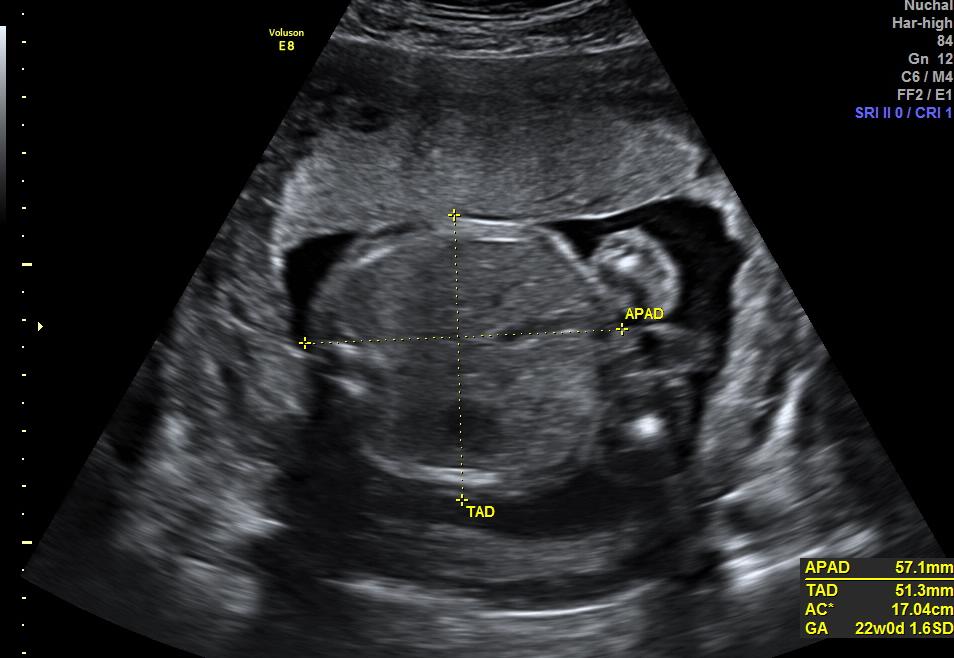}} 
	    &\parbox[c]{2.9cm}{\centering	\includegraphics[width=2.9cm]{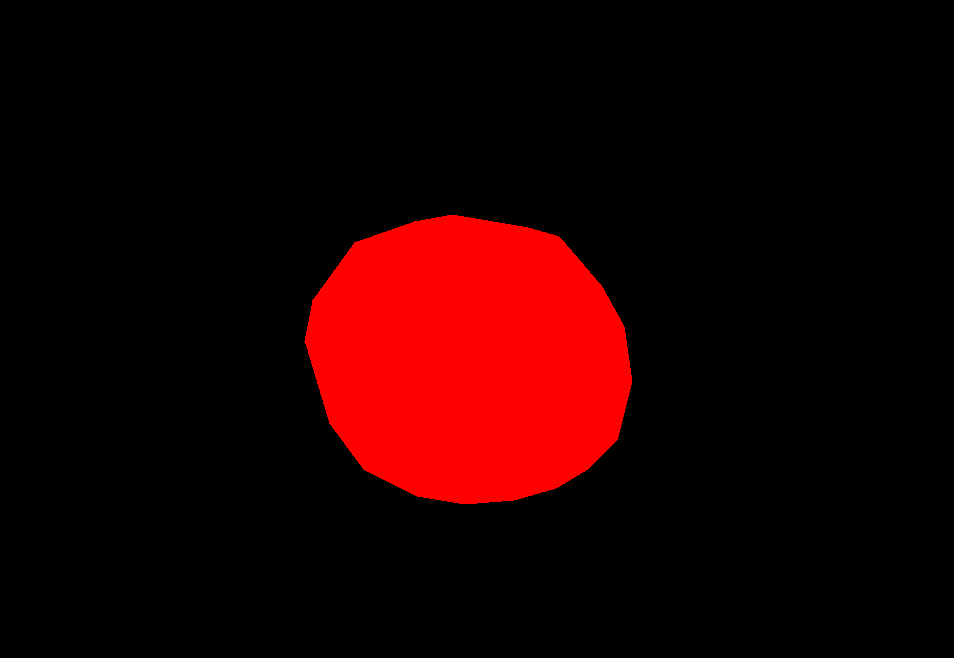}} 
	    &\parbox[c]{2.9cm}{\centering \includegraphics[width=2.9cm]{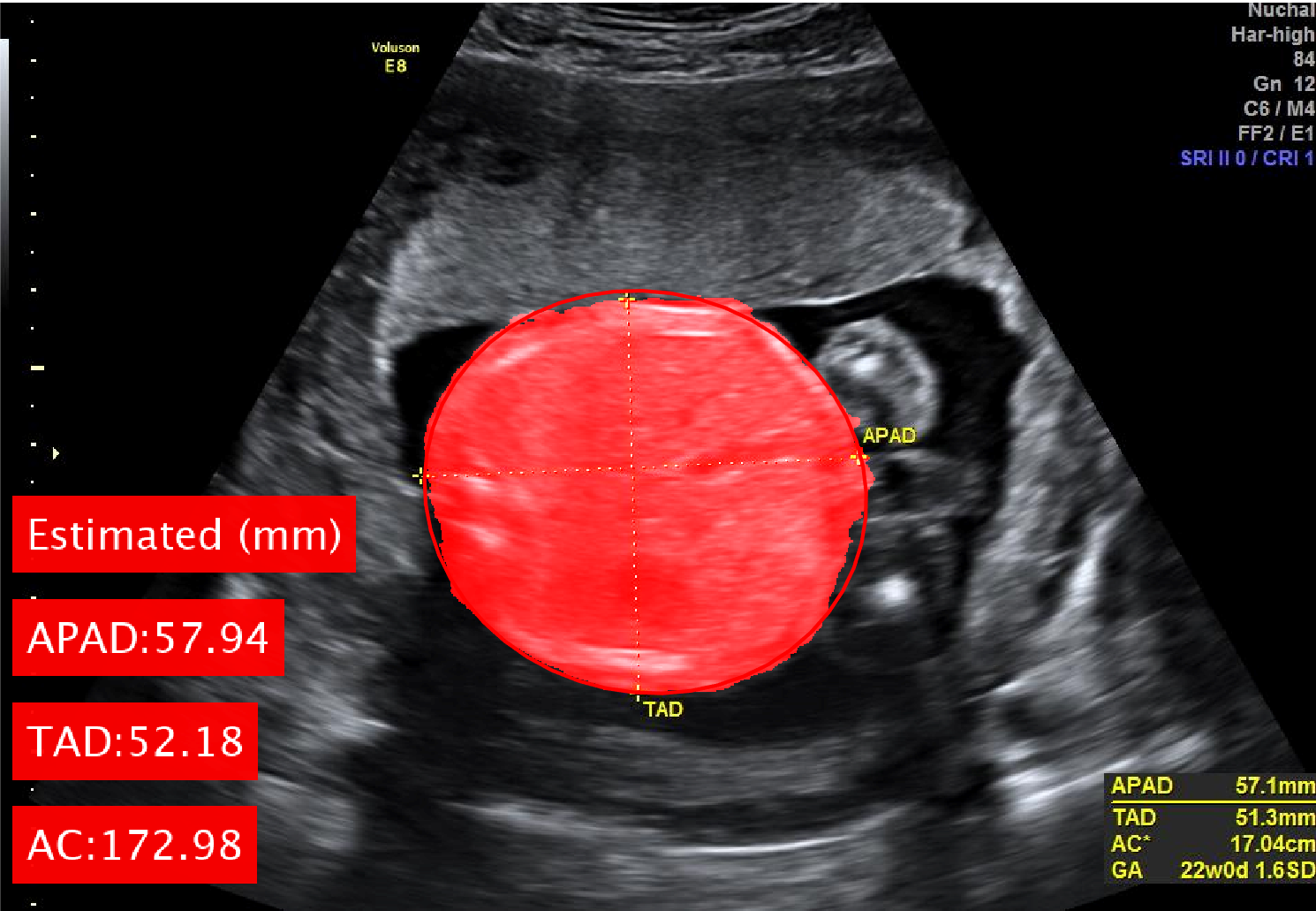}} 
	    &\parbox[c]{2.9cm}{\centering \includegraphics[width=2.9cm]{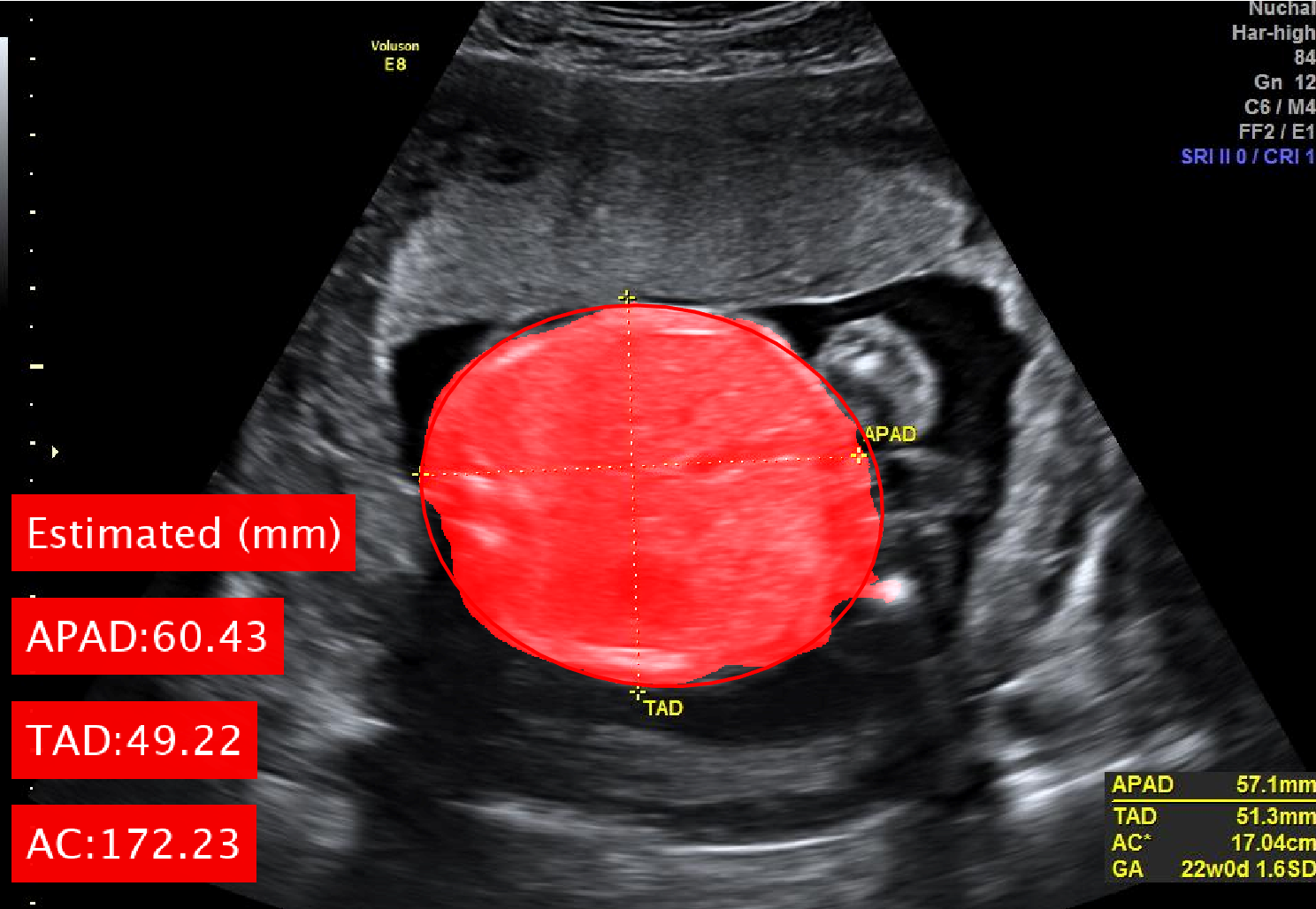}}
	    &\parbox[c]{2.9cm}{\centering \includegraphics[width=2.9cm]{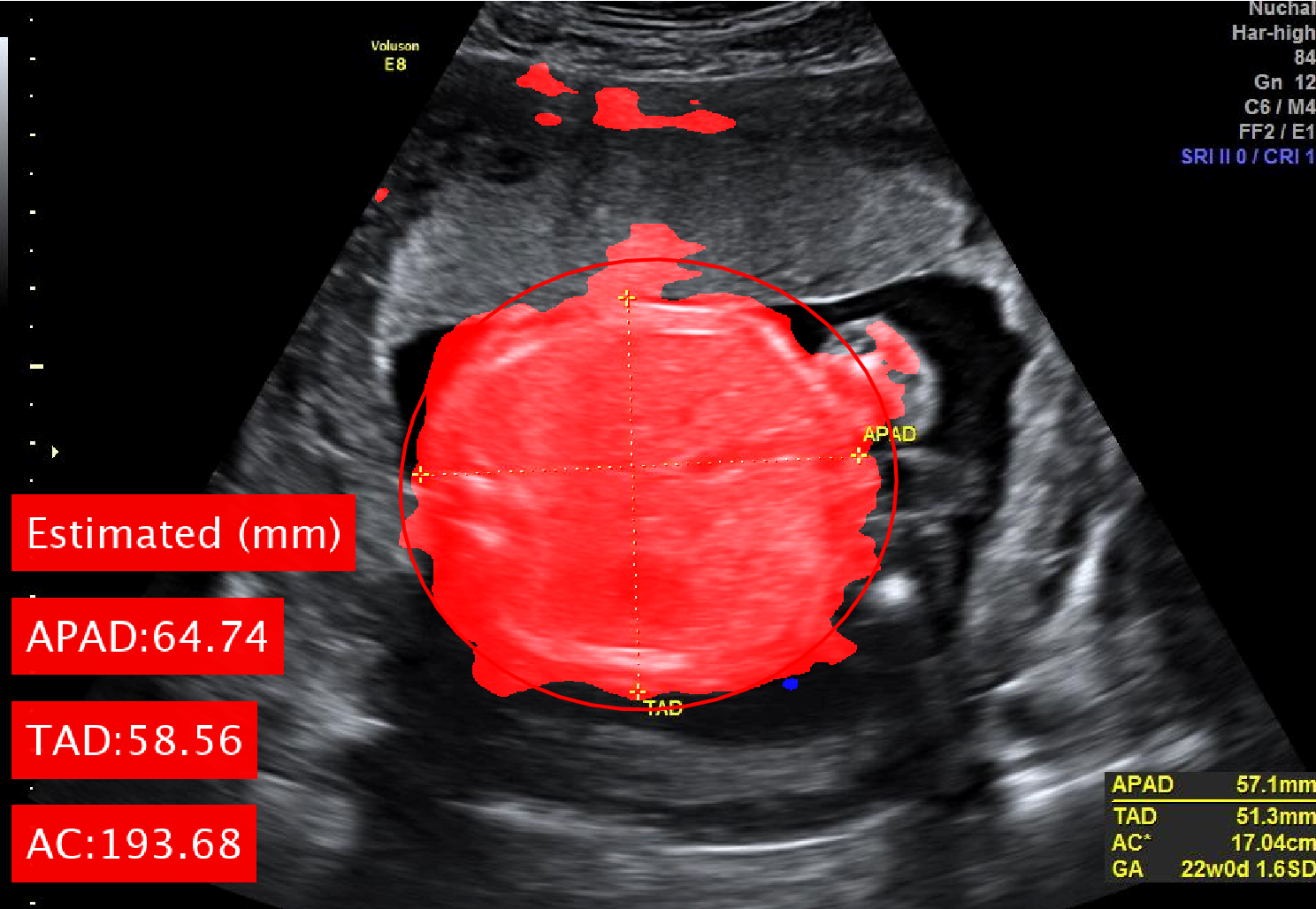}} 
	    \\
	\rotatebox[origin=c]{90}{FL} &\parbox[c]{2.9cm}{\centering \includegraphics[width=2.9cm]{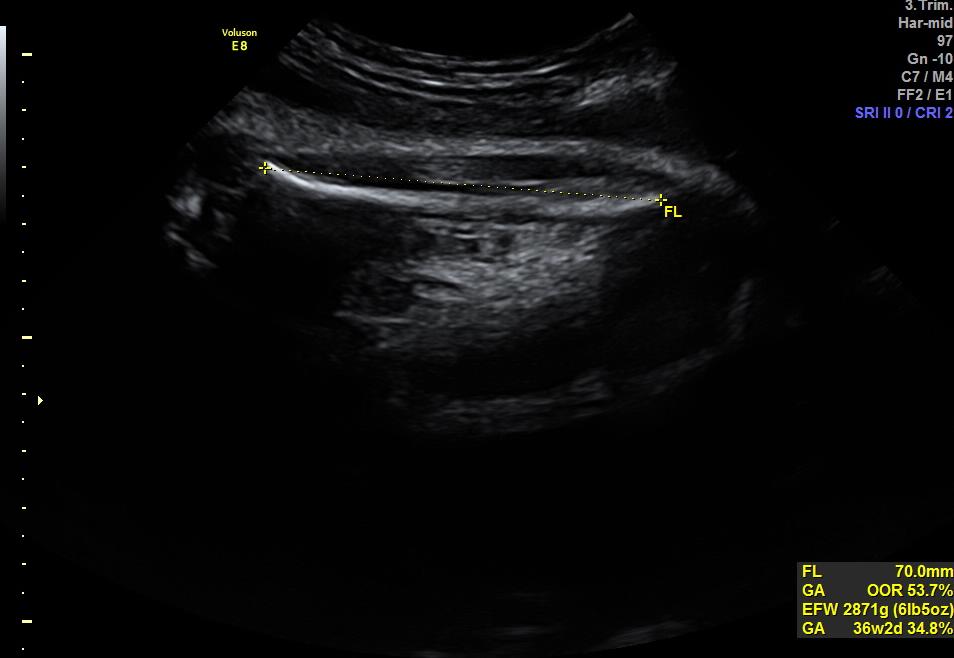}} 
	    &\parbox[c]{2.9cm}{\centering	\includegraphics[width=2.9cm]{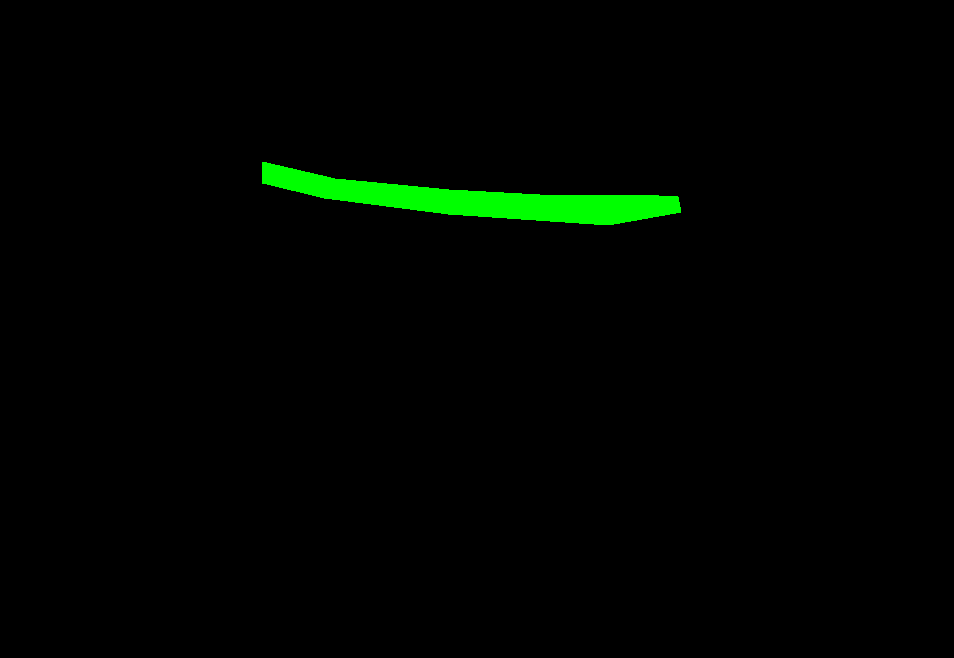}} 
	    &\parbox[c]{2.9cm}{\centering \includegraphics[width=2.9cm]{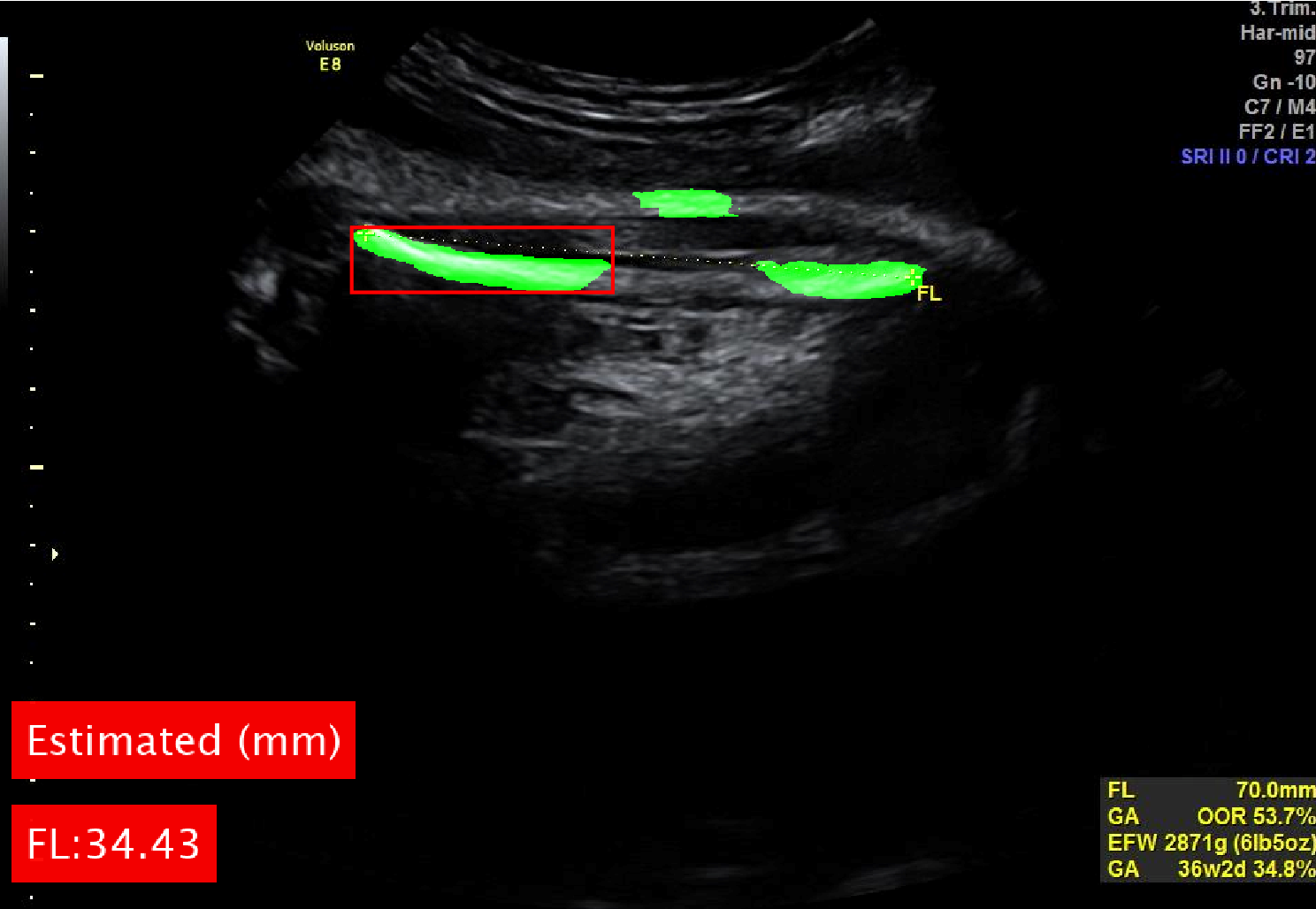}} 
	    &\parbox[c]{2.9cm}{\centering \includegraphics[width=2.9cm]{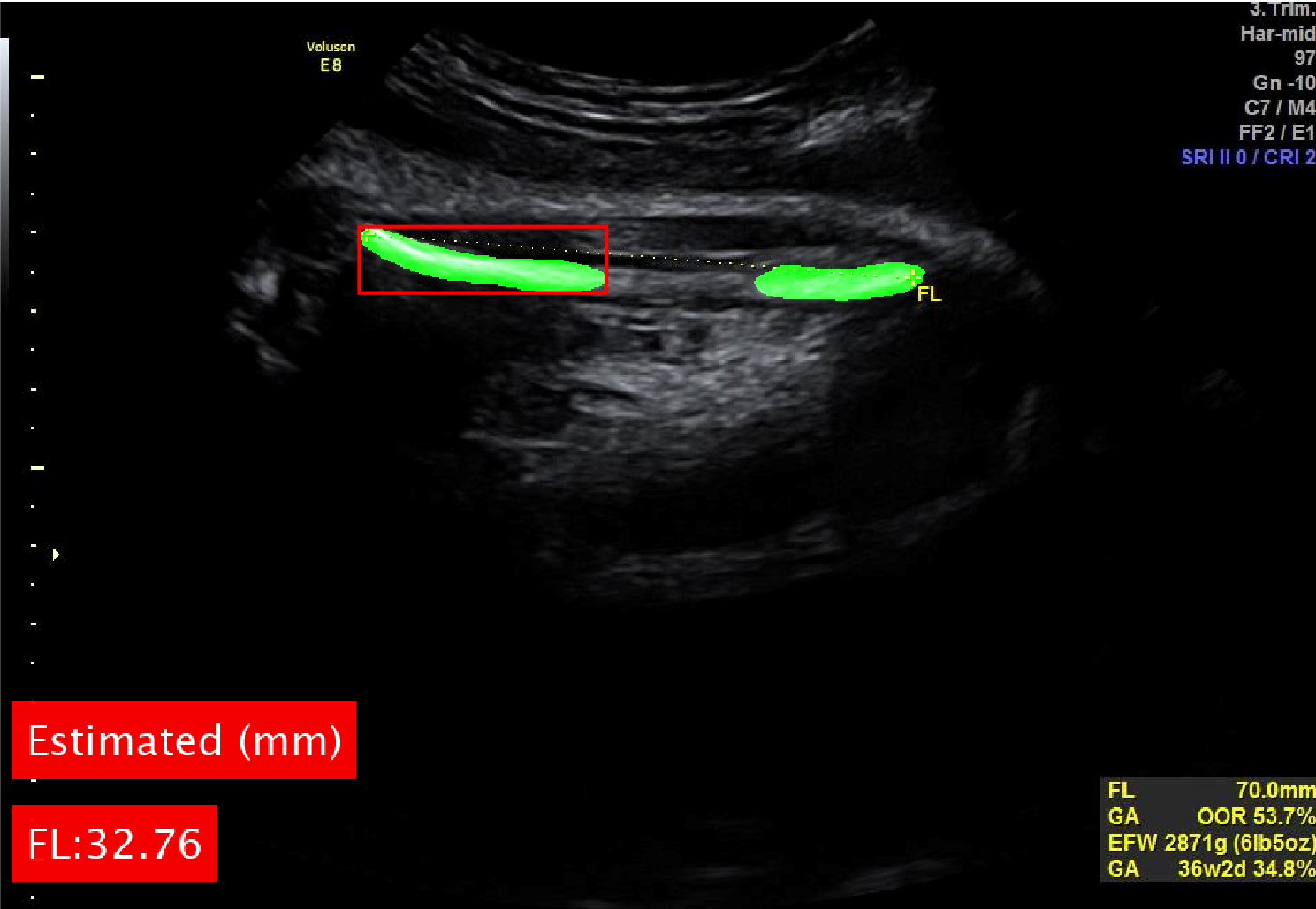}}
	    &\parbox[c]{2.9cm}{\centering \includegraphics[width=2.9cm]{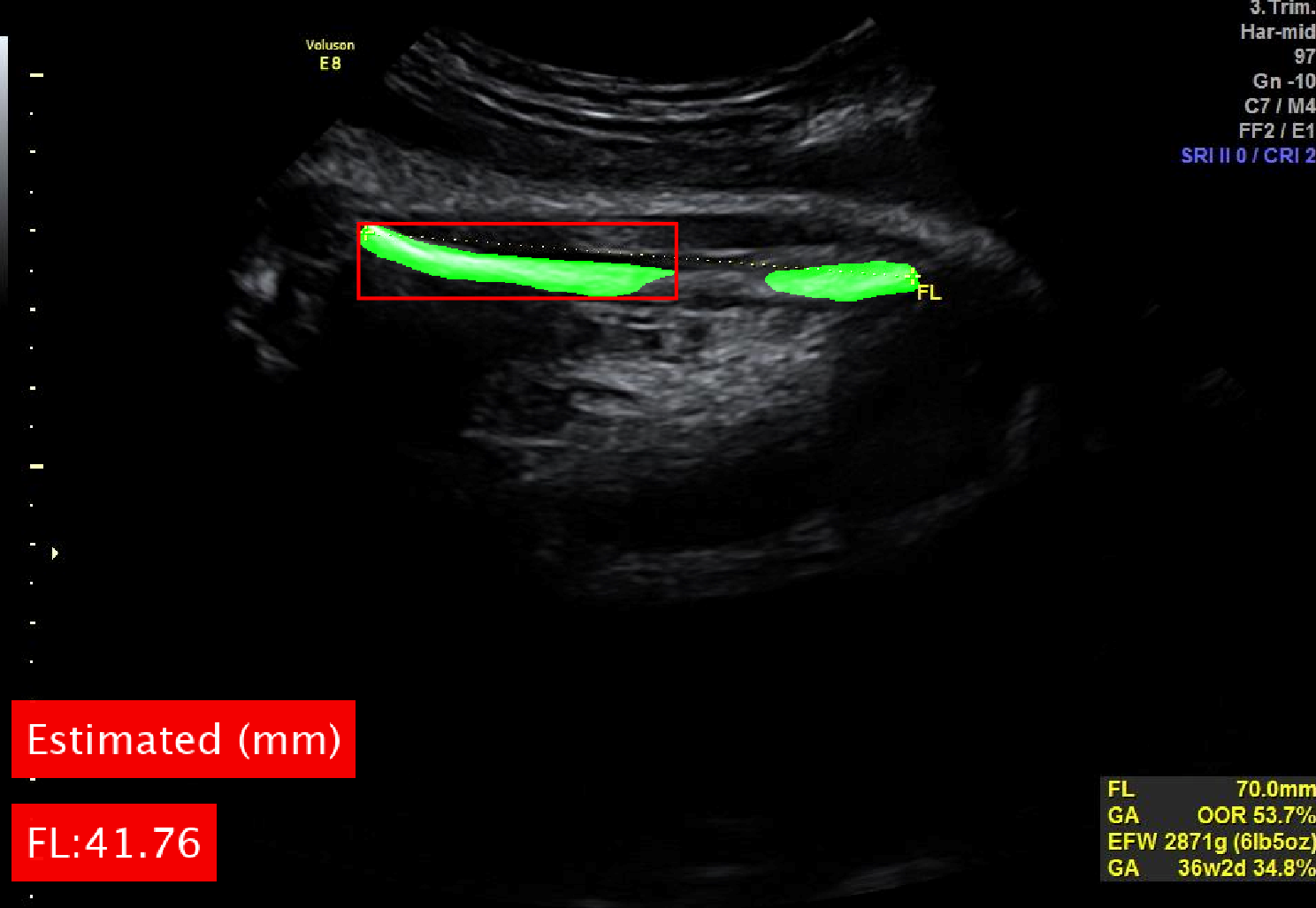}} 
	    \\
	\end{tabular}
	}
	\caption{Qualitative comparison of segmentation methods showing scenarios where inaccurate segmentation resulted in fetal biometry estimation failure. (Row 1 and 2) HC and AC examples where UNet resulted in inaccurate segmentation. (Row 3) FL example where all three methods failed. This image corresponds to the only outlier which is visible in Fig.~\ref{fig:Err_plots}(c) Deeplabv3+ error plot.} 
	\label{fig:vis_results}
\end{figure*}

Figure~\ref{fig:Err_plots} shows the boxplots for the absolute error between the clinically measured and predicted biometry. The error in head measurements are the lowest, with a median of $0.80 mm$ for BPD, $1.30 mm$ for OFD and $2.67 mm$ for HC and fewer outliers compared to other methods when segmentation masks from Deeplabv3+ (Mobilev2+wCE) are used (Fig.~\ref{fig:Err_plots}(a)). A similar trend is observed for the abdomen measurements, with a median of $2.39 mm$ for TAD, $3.82 mm$ for APAD and $3.77 mm$ for AC (Fig.~\ref{fig:Err_plots}(b)). FL showed comparable results with a median of $2.1 mm$ for Deeplabv3+ (Mobilenet-v2+wCE) but with fewer outliers (Fig.~\ref{fig:Err_plots}(b)). It is worth mentioning that the obtained error is less than the $\pm15\%$ error permissible in the US assessment~\cite{sarris2012intra}. 
Figure~\ref{fig:vis_results} presents the qualitative comparison of the segmentation methods, depicting cases where either one or all methods fail in estimating the biometry due to inaccurate segmentation.

From a clinical point of view, successful interpretation of clinical US images requires an understanding that the fetus, a 3D object, fixed in neither time nor space is being represented on a 2D grey-scale. Operator experience, combined with the effects of probe motion and homogeneity of US images contributes to high inter- and intra-operator variability. US is used extensively in the assessment and management of pregnancies at high risk of fetal growth disorders. Appropriate management of these cases requires high quality assessment and reproducible assessment of fetal weight, which can be achieved through AutoFB as demonstrated from the obtained results.

\section{Conclusion}
We proposed AutoFB, a unified framework for estimating fetal biometry given the three standard US planes. The proposed framework exploited the existing segmentation networks for predicting the segmentation masks for the head, abdomen and femur. Head and abdomen were modelled as an ellipse with their major and minor axes and circumference providing an estimate for the respective measurements. Femur length was modelled as the diagonal on a rectangle fitted onto the segmentation mask. Through retrospective scale recovery and shape fitting, we obtained the fetal biometry estimates. Comparison of the predicted versus clinically measured fetal biometry showed that the errors in HC ($2.67 mm$), AC ($3.77 mm$) and FL ($2.10 mm$) were minimal and were better than the $\pm15\%$ error that is typically acceptable in fetal US assessment. Future work involves increasing the training data size for further improving the segmentation and integrating AutoFB with the standard US plane detection~\cite{baumgartner2017sononet} framework. Moreover, comparing experts and novices performance with the AutoFB can provide evidence supporting its clinical translation.

\subsubsection{Acknowledgments}The work was supported by the Wellcome/EPSRC Centre for Interventional and Surgical Sciences (WEISS) [203145Z/16/Z]; Engineering and Physical Sciences Research Council (EPSRC) [EP/P027938/1, EP/R004080/1, EP/P012841/1, NS/A000027/1]; the H2020 FET (GA 863146); Wellcome [WT101957]; and The Royal Academy of Engineering Chair in Emerging Technologies scheme (CiET1819/2/36). A. L. David and D. M. Peebles are supported by the National Institute for Health Research University College London Hospitals Biomedical Research Centre.

%
\bibliographystyle{splncs04}
\bibliography{references}

\end{document}


\title{AutoFB: Automating Fetal Biometry Estimation from Standard Ultrasound Planes
(Supplementary Material)}
\titlerunning{AutoFB: Automating Fetal Biometry Estimation}


\author{Sophia Bano \inst{1,2} \and
Brian Dromey \inst{1,3} \and 
Francisco Vasconcelos \inst{1,2} \and
Raffaele Napolitano \inst{3} \and 
Anna L. David \inst{3} \and 
Donald M Peebles \inst{3} \and
Danail Stoyanov \inst{1,2}}

\institute{Wellcome/EPSRC Centre for Interventional and Surgical Sciences(WEISS), University College London, London, UK  \\
Department of Computer Science, University College London, London, UK \and
Elizabeth Garrett Anderson Institute for Women's Health, University College London, London, UK}

\authorrunning{S. Bano et al.}

\maketitle

\begin{figure}[h!]
\includegraphics[width=0.32\textwidth]{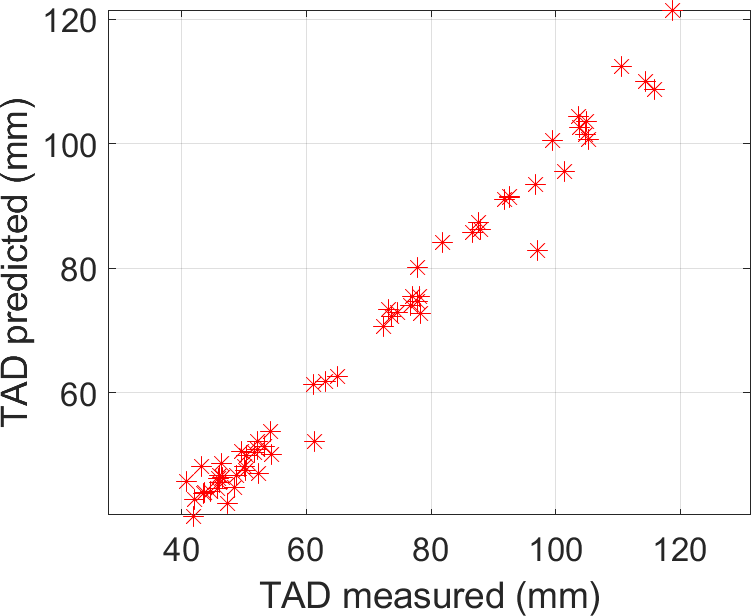}
\includegraphics[width=0.32\textwidth]{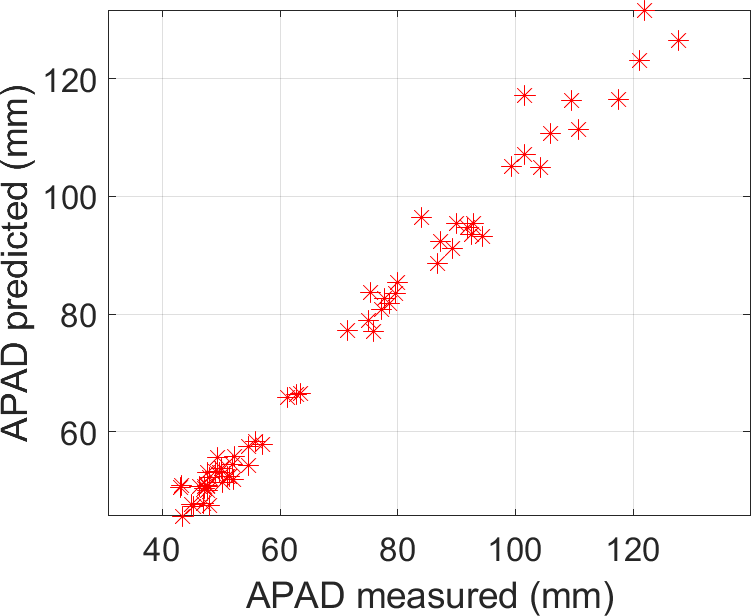}
\includegraphics[width=0.32\textwidth]{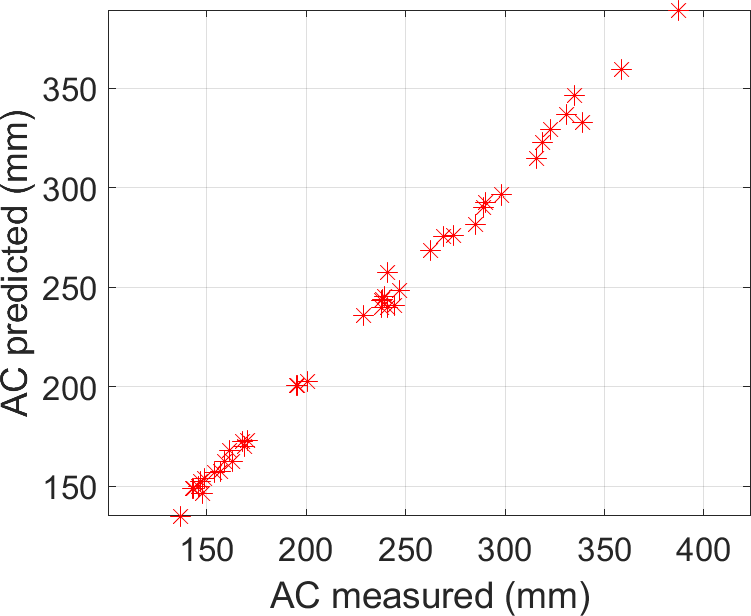}
\\
\includegraphics[width=0.32\textwidth]{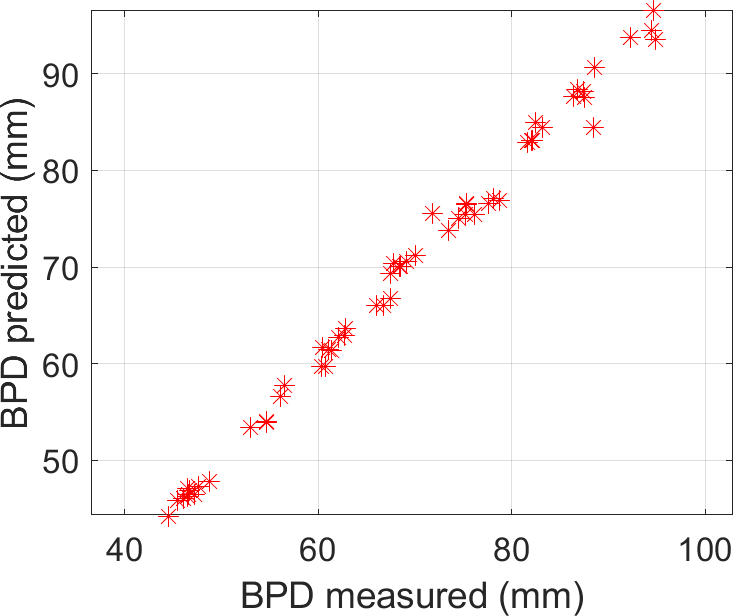}
\includegraphics[width=0.32\textwidth]{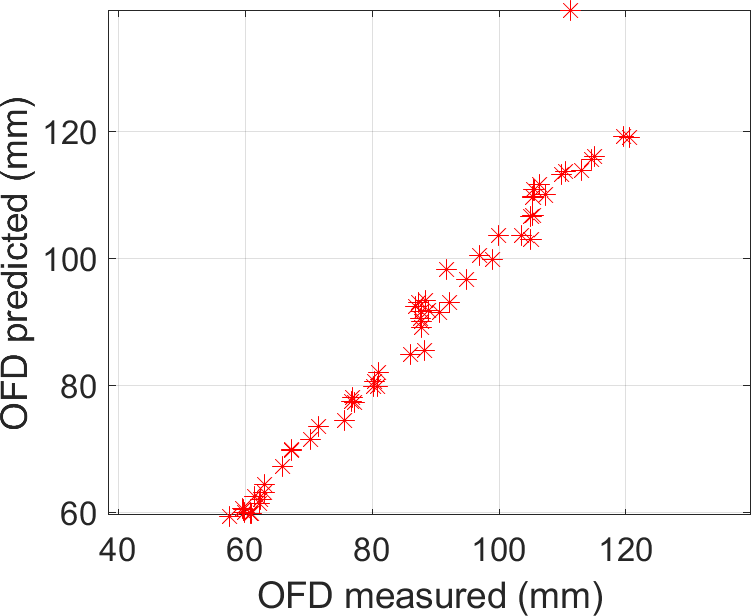}
\includegraphics[width=0.32\textwidth]{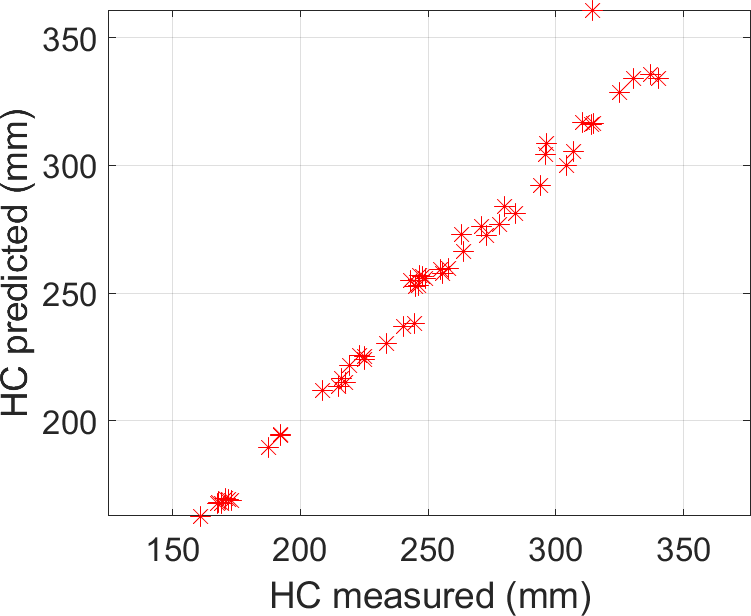}
\\
\centering
\includegraphics[width=0.32\textwidth]{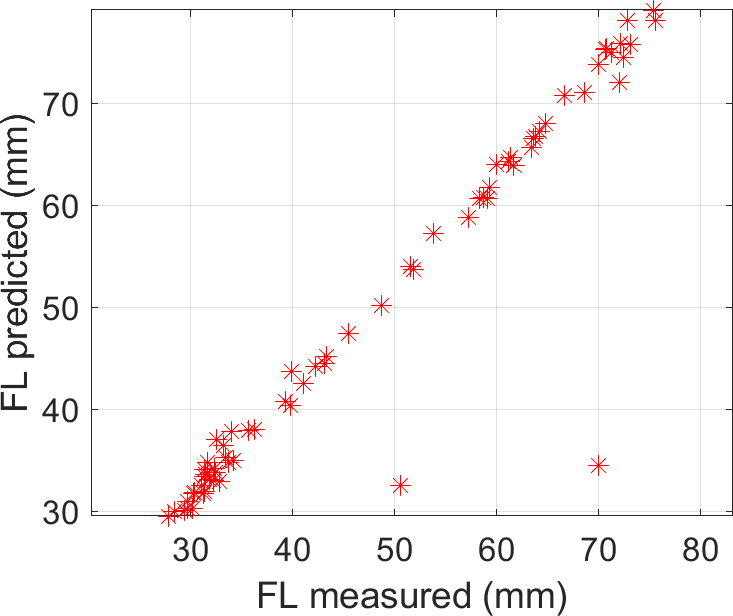}
\caption{Predicted fetal biometry from the best performing architecture (Deeplabv3+ with MobileNetv2) as reported in Table 2 and Fig. 3 versus clinically measured fetal biometry is shown plotted for TAD, APAD, AC, BPD, OFD, HC and FL. Observe that all measurements lie on a diagonal with a few outliers evident in OFD, HC and FL.}
\label{fig:Pred_vs_Measured}
\end{figure}